# Multi-Level Analysis and Annotation of Arabic Corpora for Text-to-Sign Language MT

**Abdelaziz Lakhfif**
Informatique Department, Ferhat Abbas University, Sétif, Algeria
`aziz_lakhfif@yahoo.fr`

**Mohammed T. Laskri**
LRI-GRIA, Badji Mokhtar University, BP 12 Annaba Algeria
`laskri@univ-annaba.org`

**Eric Atwell**
School of Computing, University of Leeds, Leeds, LS2 9JT, UK
`eric@comp.leeds.ac.uk`

## 1  Introduction

The Arabic language is morphologically rich and syntactically complex with many differences from European languages, and this creates a challenge when porting existing annotation tools to Arabic.

In this paper, we present an ongoing effort in lexical semantic analysis and annotation of Modern Standard Arabic (MSA) text, a semi automatic annotation tool concerned with the morphologic, syntactic, and semantic levels of description. Besides the aim of providing a multi-level annotation tool for Arabic corpora, our goals are (1) to investigate the suitability of Frame Semantics (FS) approach (Fillmore 1985) for representing and analysing Arabic text (2) to provide corpus-attested linguistics materials for frame-based contrastive text analysis between Arabic and English in terms of lexicalization patterns; (3) to automatically derive mappings rules from annotated sentences. Such corpus-attested mapping rules between linguistic form and its meaning can support semantic analysis in knowledge-based NLP systems such as machine translation, information extraction etc.

Following syntactically-based annotation projects for English, serious attempts have been made to annotate Arabic corpora, such as the Penn Arabic Treebank (PATB), (Maamouri et al. 2004) and the Quranic Arabic Dependency Treebank (QADT) (Dukes et al. 2010). However, semantically-based annotation for Arabic corpora has not yet been garnering the same attention.

Our semantic representations are based upon use of frame-semantic paradigm; it is actually used in a MT system from Arabic to Algerian Sign Language aimed to assist deaf children and in order to bridge the gap between Arabic written texts and Algerian Sign Language (Lakhfif and Laskri 2010a,b, 2011).

Annotation outputs are available in XML format compatible with the FrameNet project (Fillmore and Petruck 2003) design and can be portable to other NLP systems.

## 2  Resources

During the last decade, the emergence of semantically rich lexical resources has been one of the most striking advances in the NLP domain.

The annotation also make use of publicly existing lexical semantic resources such as Buckwalter Arabic Morphological analyzer (BAMA) (Buckwalter 2002) lexicon, FrameNet (Baker et al. 1998), and Arabic WordNet (AWN) (Elkateb et al. 2006), in order to facilitate and scale up to wide-coverage annotation tasks.

FrameNet project is the most important online semantic lexicons resource for the English language, based on Fillmore's Frame Semantics tenet. The database is validated with semantically and syntactically annotated texts from real linguistic data. The success of the Berkeley FrameNet for English motivates similar projects to start producing comparable frame-semantic lexicons for other language such as German, Japanese, Spanish, Danish, Swedish, etc. (Boas 2009). For our project, we are re-using frames from the Berkeley FrameNet project for English to construct an equivalent FrameNet for Arabic following the process described in (Fillmore and Petruck 2003). We have also used AWN to extend the FrameNet coverage for Arabic.

AWN is a lexical database of Arabic words in terms of a large semantic network. AWN follows the methodology developed for EuroWordNet and it was mapped to Princeton WordNet 2.0 and SUMO ontology.

BAMA is a rule based morphological analyzer that provides for every lexicon entry, a morphological compatibility category, an English gloss and occasional part-of-speech (POS) tags.

## 3 Arabic corpus and annotation process

In order to cover linguistic variation of the language, we aim to use several Arabic corpora from different sources such as Quranic texts, and CCA Corpus of Contemporary Arabic (Al-Sulaiti and Atwell 2006). We have started with a collection of Arabic texts from Algerian primary school educational books, actually used as a development corpus in our Arabic text-to-Sign Language MT system for Algerian deaf children. As regard sentences extraction from corpora, we follow FrameNet Project data corpus collection and organisation. So for each word target, sentences containing the word with the appropriate sense are extracted from a different corpus sources and classified into sub-corpora by syntactic pattern. Each annotation level consists of some label sets arranged in layers of annotation such as "**FE**" for Frame Element annotation, "**GF**" for Grammatical Function annotation, "**Sumo**" for Sumo concept annotation, etc. Data annotations are organized by Lexical Unit, that is, one file contains data for all layers. Also, the tool can provide separated data annotation for selective layers and it can be performed in incremental fashion. In the pilot phase of the annotation, more than 1,000 Arabic sentences expressing Motion events were annotated with our tool.

### 3.1 Relevant linguistic features for Arabic

Besides diacritics to mark the syntactic cases, Arabic encodes nominal categories with inflectional features like gender, number and definiteness. The verb encodes gender, number, person, tense, voice and mood features. Syntactically, Arabic is a pro-drop language. That is, its verbs incorporate their subject pronouns as part of their inflection. Also, in some constructions, the object can be encoded in the verb morphology. For example (table 1) the word (fa>akalotuhA) consists of the conjunction "ف-fa" "and/so", a verb "أكل->akala-eat" inflected for the masculine singular first person subject pronoun in the past tense and a feminine singular 3rd person object pronoun.

| Arabic: | فَأَكَلْتُهَا |
|---|---|
| Transliteration : | fa>akalo--tu--hA |
| POS : | Conj--V.$_{pass}$--Pro.$_{1sm}$(subj)--Pro.$_{3sf}$(obj) |
| Gloss: | and--Eat--I--them   (and I eat them) |

Table 1: Arabic morphology complexity

Morpho-syntactic features are useful information in disambiguation tasks in morphologically rich languages like Arabic. Further, word pattern, derivational information (verbal noun-مصدر , active participle- اسم الفاعل, etc.), and predicate-arguments structure also seem to be very important in some cases in which morphological features alone cannot predict the appropriate syntactic relation. Finally, semantic information about arguments are necessary to disambiguate accusative NPs (objects) in di-transitive constructions.

### 3.2 Lexico-morphological annotation level

Lexico-morphological annotation starts with word segmentations (some lexical entries are the concatenation of two or three word segments with different POS tags) in terms of their POS tags and aims to provide for the main segment word all its possible lexical items with their properties such as stem, root, and additional specific morpho-syntactic features.

In order to perform a fine grained analysis of Arabic texts, we integrate the BAMA analyzer with a lexical semantics analyzer based on AWN lexicons. Since BAMA like most available Arabic morphology-based analyzers does not provide semantic information, AWN-based analysis improves tokens with lexical semantics, which has proved helpful to bridge the semantic gap and the coverage issue by using its rich network of semantic relations between words.

### 3.3 Syntactic annotation level

Arabic can be seen as a language that has a network of dependency relations in every phrase or clause (Ryding 2005). Thus, syntactic dependency analysis is very suitable for free word-order languages such as Arabic. Since our aim is to capture the features of meaning anchored in features of linguistic form, dependency-based syntactic analysis seems more helpful for semantic analysis. As a richer set of function labels improves the argument classification task, so, we use additional detailed relevant relations such as adjective (صفة-Sifa), cognate accusative (المفعول المطلق), adverbial accusative of cause (المفعول لأجله), etc. The process is based upon use of a set of agreement features including morphological, syntactic, and semantic information. In addition to dependency annotation scheme, we have extended the syntactic annotation with constituents based annotation compatible with FrameNet annotation scheme.

### 3.4 Frame-Semantics annotation level

At the semantic annotation level, our strategy rests on the multilingual dimension of the Frame Semantics approach. Following FrameNet annotation procedures, the annotation process realizes a mapping between linguistics forms and cognitive structures. Annotation steps consists of (a) identifying the predicate and their dependent arguments in the sentence. The predicate should be recognized as target word that evokes the frame, (b) selecting the appropriate semantic frame describing the event or the situation using the frame definition from the FrameNet Database, (c) selecting for each argument describing the participants in the event, its corresponding semantic role (FE). In order to facilitate the annotation task, the tool provides samples of annotated sentences from English FrameNet.

### 4 Data representation and exploitation

Our tool uses an XML-based format to represent data corpus and annotation information, from which, we are deriving linguistic generalization (figure1.) from different annotation levels as mapping rules. Besides all possible patterns in the configurations of grammatical function (GF), frame elements (FE) and phrase type (PT) information that characterizes each lexical unit (LU), derived rules contain the semantic field of the target and some other morpho-syntactic features that support disambiguation of Arabic texts.

```
<m id="01" pattern="VP.NP-nom.NP-acc.PP(في)" voice="A" cons="T">
    <rule id="011" FE ="Agent" PT="NP-nom" GF="Subj"/>
    <rule id="012" FE ="Theme" PT="NP-acc" GF="Obj"/>
    <rule id="013" FE ="Source" PT="PP-gen" GF="OBL"/>
</m>
```

Figure. 1: rules from Placing Frame annotation.

### 5 Conclusion and future work

We presented a multi level annotation project, the aim of which is to analyse and annotate Arabic corpora with morpho-syntactic and semantic information.

In our first step, our focus is on annotation of Arabic texts collected from first level educational books, in order to automatically generate mapping rules that could be used in our Arabic text-to-sign Language MT for Algerian Deaf children. The annotation corpus can be used as data source for Arabic NLP systems. In future, we plan to release an enhanced version of our tool for online annotation of large Arabic corpus.

## Appendix

Figure 2. a screenshot of a sample illustrating the data format of the Arabic corpus.

Figure 3. A screenshot of the annotation Tool and a sample XML-based output (FS annotation)